\documentclass[letterpaper]{article}

\usepackage{ijcai19}
\usepackage{amssymb}
\usepackage[english]{babel}
\usepackage{times}
\usepackage{helvet}
\usepackage{courier}
\usepackage{graphicx}
\usepackage{comment}
\usepackage{booktabs}
\usepackage{dblfloatfix}

\title{Topological Data Analysis for Arrhythmia Detection \\ through Modular Neural Networks}
\author{Meryll Dindin$^{[1]}$, Yuhei Umeda$^{[2]}$, Frederic Chazal$^{[3]}$ \\ \vspace{0.1in} $^{[1]}${\small meryll.dindin@student.ecp.fr} $^{[2]}${\small umeda.yuhei@jp.fujitsu.com} $^{[3]}${\small frederic.chazal@inria.fr}}

\begin{document}

\maketitle

\begin{abstract}
    This paper presents an innovative and generic deep learning approach to monitor heart conditions from ECG signals.We focus our attention on both the detection and classification of abnormal heartbeats, known as arrhythmia. We strongly insist on generalization throughout the construction of a deep-learning model that turns out to be effective for new unseen patient. The novelty of our approach relies on the use of topological data analysis as basis of our multichannel architecture, to diminish the bias due to individual differences. We show that our structure reaches the performances of the state-of-the-art methods regarding arrhythmia detection and classification.
\end{abstract}

\section{Introduction}

\indent Countless artificial intelligence breakthroughs are observed in healthcare on a daily basis. They currently target improved monitoring of vital signs, better diagnostics and more reliable clinical decisions. Among the many on-going developments, heart monitoring is of particular importance as heart attack and strokes are among the five first causes of death in the US. Developing wearable medical devices would help to deal with a larger proportion of the population, and reduce the time used by cardiologists to make their diagnosis. This paper focuses on both the detection and classification of arrhythmia, which is an umbrella term for group of conditions describing irregular heartbeats. Detection deals with spotting any abnormal heartbeat, while classification deals with giving the right label to the spotted abnormal heartbeats.

Among the several existing studies, some developed descriptive temporal features to feed SVM \cite{Houssein} or neural networks \cite{SHADMAND201612}, sometimes mixed with optimization methods \cite{Houssein,ABDELAZIM2013334}. Others \cite{YOCHUM201646} dealt with wavelet transforms and Daubechies wavelets. The general approach of those papers enables arrhythmia classification through machine learning. However, most papers \cite{KORA201644,Jun,Li2018,Hassanien2018CombiningSV} reduce the classification to specific arrhythmia, or limit it to a few classes only. On the other hand, \cite{Rajpurkar} sought to improve multi-class classification. But the more the classes, the faster the performances did vanish. To overcome this issue, \cite{Li2018,Clifford} have introduced deep learning methods based on convolutional networks. Other teams focused on unsupervised learning, such as auto-encoders \cite{Weijia}, with promising results. Nonetheless, the methods presented so far have low performance for unknown patient, mainly due to individual differences. Generalization, which means independence on individual differences, is a serious issue for any application in the healthcare sector.

The proposed approach consists in the analysis of ECG through a modular multi-channel neural network whose originality is to include  a new channel relying on the theory of topological data analysis, that is able to capture robust topological patterns of the ECG signals. That information describes best the geometry of each heartbeat, independently of the values of the signal or the individual heart rhythms. By combining topological data analysis, handcrafted features and deep-learning, we aimed for better generalization. 

Our paper is organized as follows. After presenting Topological Data Analysis, we condensed our approach in the presentation of the datasets, our preprocessing and the general deep-learning architecture. We then develop our testing methodology, which is used to quantify generalization. The last sections provide comparisons with benchmarks and state-of-the-art results, and conclude with our experimental results. We introduce a new benchmark for arrhythmia classification, underlining the strengths of topological data analysis and auto-encoders to tackle the issue of individual differences. Finally, remarks and thoughts are provided as conclusion at the end of the paper.

\begin{figure*}[hb!]
\centering
\includegraphics[width=0.95\textwidth]{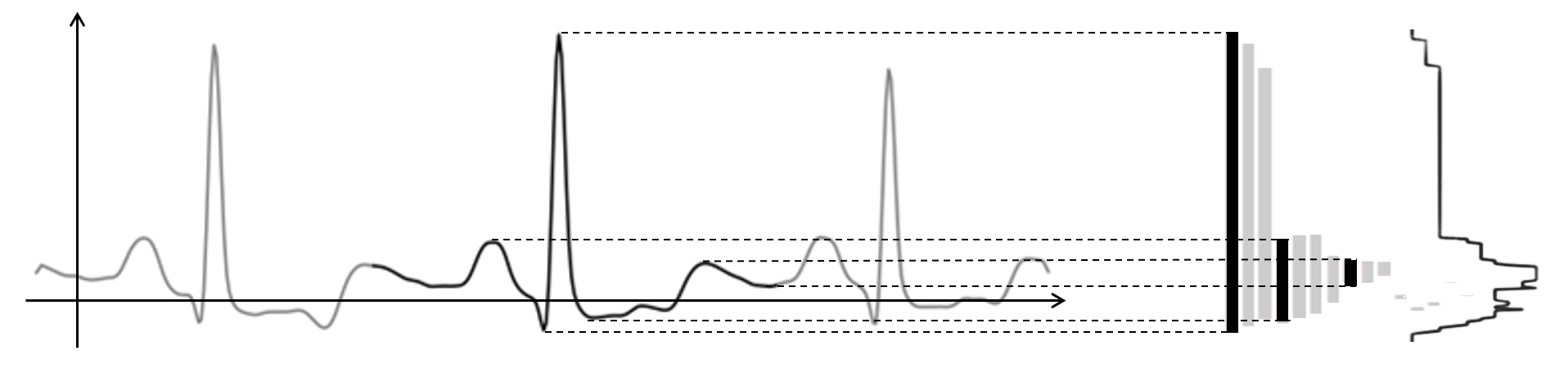}
\caption{From three Consecutive Heartbeats to their corresponding persistence barcode and Betti curve}
\label{figure:ecg_to_betti}
\end{figure*}

\section{Topological Data Analysis}

Among the main challenges faced for arrhythmia classification generalization, we find individual differences, and specifically bradycardia and tachycardia. We dealt with it by introducing Topological Data Analysis, and by merging theory with our deep-learning approach. Topological Data Analysis (TDA) is a recent and fast growing field that provides mathematically well-founded methods \cite{Chazal} to efficiently exhibit topological patterns in data and to encode them into quantitative and qualitative features. In our setting, TDA, and more precisely {\em persistent homology theory} \cite{edelsbrunner2010computational}, 
powerfully characterizes the shape of the ECG signals in a compact way, avoiding complex geometric feature engineering. Thanks to fundamental stability properties of persistent homology \cite{chazal2012structure}, the TDA features appear to be very robust to the deformations of the patterns of interest in the ECG signal, especially expansion and contraction in the time axis direction. This makes them particularly useful to overcome individual differences and potential issues raised by bradycardia and tachycardia.

\begin{figure}[ht!]
\centering
\includegraphics[width=0.45\textwidth]{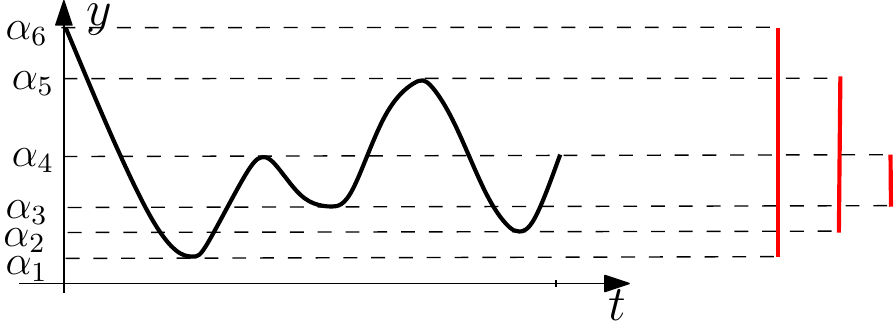}
\caption{The persistence barcode of a 1D signal}
\label{figure:PeristFn1D}
\end{figure}

\paragraph{Persistence Homology.} To characterize the heartbeats, we consider the persistent homology of the so called sub-level (resp. upper-level) sets filtration of the considered time series. Seeing the signal as a function $f$ defined on an interval $I$ and given a threshold value $\alpha$, we consider the connected components of $F_\alpha = \{ t \in I: f(t) \leq \alpha \}$ (resp. $F^\alpha = \{ t \in I: f(t) \geq \alpha \}$). As $\alpha$ increases (resp. decreases) some components appear and some others get merged together. Persistent homology keeps track of the evolution of these components and encodes it in a {\em persistence barcode}, i.e. a set of intervals - see Figure \ref{figure:PeristFn1D} for an example of barcode computation on a simple example. The starting point of each interval corresponds to a value $\alpha$ where a new component is created while the end point corresponds to the value $\alpha'$ where the created component gets merged into another one. In our practical setting, the function $f$ is the piecewise linear interpolation of the ECG time series and persistence barcodes can be efficiently computed in $O(n \log n)$ time, using, e.g., the GUDHI library \cite{maria2014gudhi}, where $n$ is the number of nodes of the time series.

\begin{figure*}[hb!]
\centering
\includegraphics[width=0.95\textwidth]{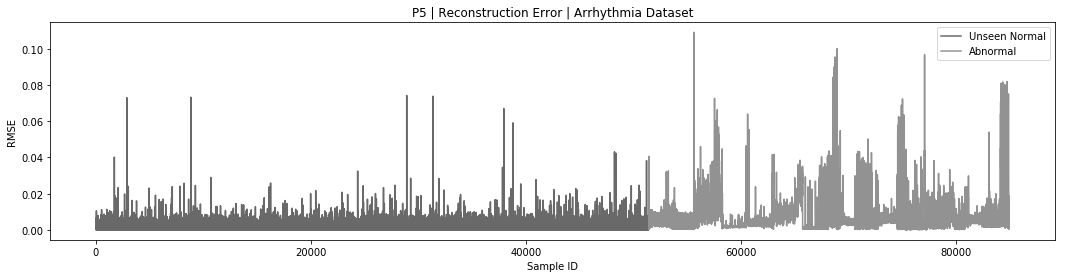}
\caption{Reconstruction Error on Heartbeat Signals}
\label{figure:reconstruction_error}
\end{figure*}

To clarify the construction of a persistence barcode, one may observe Figure \ref{figure:PeristFn1D} with the following notations: $y=f(t)$: for $\alpha < \alpha_1$, $F_\alpha$ is empty. A first component appears in $F_\alpha$ as $\alpha$ reaches $\alpha_1$, resulting in the beginning of an interval. Similarly when $\alpha$ reaches $\alpha_2$ and then $\alpha_3$, new components appear in $F_\alpha$ giving birth to the starting point of new intervals. When $\alpha$ reaches $\alpha_4$, the two components born at $\alpha_1$ and $\alpha_3$ get merged, resulting in the ``death'' of the most recently born component (persistence rule), i.e. the one that appeared at $\alpha_3$ and creation of the interval $[\alpha_3,\alpha_4]$ in the persistence barcode. Similarly when $\alpha$ reaches $\alpha_5$ the interval $[\alpha_2,\alpha_5]$ is added to the barcode. The component appeared at $\alpha_1$ remains until the end of the sweeping-up process, resulting in the interval $[\alpha_1,\alpha_6]$.

\paragraph{Betti Curves.} As an unstructured set of intervals, the persistence barcodes are not suitable for direct integration in machine-learning models. To tackle this issue, we use a specific representation of the barcode diagrams, the so-called Betti curves \cite{Umeda}: for each $\alpha$, the Betti curve value at $\alpha$ is defined as the number of intervals containing the value $\alpha$. The Betti curves are computed and discretized on the interval delimited by the minimum and maximum of the birth and death values of each persistent diagram, both for the time-series and its opposite (in order to study the sub-levels and upper-levels of the signal). One may observe that a fundamental property of Betti curves of 1D signal that follows from the definition of barcodes is their stability with respect to time re-parametrization and signal value rescaling, as stated in the following theorem. This allows us to build an uniform input for classical 1D convolutional deep-learning models, thus tacking the main issue of individual differences.

\paragraph{Theorem:} {\it Time Independence of Betti Curves} \\

\begin{minipage}[5cm]{0.4\textwidth}
    Given a function $ f : I \to \mathbb{R}$ and a real number $a>0$ the Betti curves of $t \to f(t)$ and $t \to f(at)$ are the same. \\ \\
    Moreover, if $g(t) = b f(t)$ for some $b >0$, then the Betti curves of $f$ and $g$ are related by $BC_g(\alpha) = BC_f(\frac{\alpha}{b})$.\\
\end{minipage}

This theorem is a particular case of a more general statement resulting from classical properties of general persistence theory \cite{chazal2012structure}. Intuitively, the invariance to time rescaling follows from the observation that persistence intervals measure the relative height of the peaks of the signal and not their width. The value-rescaling of the signal by a factor $b$ results in a stretching of the persistence  intervals by the same factor resulting in the above relation between the Betti curves of the signal and its rescaled version.

\section{Deep-Learning Approach}

\subsection{Datasets} 

To facilitate comparison to other existing methods, our approach is experimented on a family of open-source data sets that have already been studied among the literature. Those are provided by the Physionet platform, and named after the diseases they describe: \textit{MIT-BIH Normal Sinus Rhythm Database} \cite{Physionet}, \textit{MIT-BIH Arrhythmia Database} \cite{Physionet,Arrhythmia}, \textit{MIT-BIH Supraventricular Arrhythmia Database} \cite{Physionet,SupraVentricular}, \textit{MIT-BIH Malignant Ventricular Arrhythmia Database} \cite{Physionet,Malignant} and \textit{MIT-BIH Long Term Database} \cite{Physionet}. Those databases present single-channel ECGs, each sampled at 360 Hz with 11-bit resolution over a 10 mV range. Two or more cardiologists independently annotated each record, whose disagreements were resolved to obtain the reference annotations for each beat in the databases. Each heartbeat is annotated independently, making peak detection thus unnecessary.

\begin{table}[ht!]
{\footnotesize {\centerline{\begin{tabular}{r|c|c|c}
\toprule
\textbf{Database} & \# Patients & \# Labels & Duration (Hours) \\
\midrule
Arrhythmia & 48 & 109494 & 24 \\
LongTerm & 7 & 668735 & 147 \\
NormalSinus & 18 & 1729629 & 437 \\
SupraVentricular & 78 & 182165 & 38 \\
TWave & 90 & 790565 & 180\\
\bottomrule
\end{tabular}}}}
\caption{Physionet Datasets Description}
\label{table:dtb_description}
\end{table}

\subsection{Preprocessing}

Every machine-learning comes with its data preprocessing. We first focused on the standardization of all the available ECG. Different methods have been applied in order to enhance the signal, and reduce noise and bias. After resampling at 200 Hz, we removed the baselines \cite{Blanco-Velasco2008} and applied filters, based on both a FIR filter and a Kalman filter. The signal is then rescaled between 0 and 1 before being translated to get a mean of the signal close to 0, for deep-learning stability.

\paragraph{Baseline Wander.} The method dealing with the baseline drift \cite{Blanco-Velasco2008} is based on the Daubechies wavelets theory. It consists in consecutive processes of decomposition and reconstruction of the signal thanks to convolution windows. By removing the outlying components, we are able to identify and suppress the influence of the baseline in the signal, which generally corresponds to muscular and respiratory artifacts.

\paragraph{Filtering.} The first applied filter to each ECG is a FIR (Finite Impulse Response) filter. It performs particularly well on ECG, and wavelets-based signals in general. It behaves basically as a band-filter. We chose 0.05 Hz and 50 Hz as cut frequencies to minimize the resulting distortion, according to our tests and the literature \cite{Buendia-Fuentes2012,Upganlawar2014,Goras}.

\paragraph{Heartbeats Slicing.} Once preprocessed, each ECG is segmented into partially overlapping elementary sequences made of a fix number of consecutive heartbeats. Each sequence is extracted according to the previous and next heartbeat. This extraction being patient-dependent, it reduces the influence of diverging heartbeat rhythms, {\it e.g} bradycardia and tachycardia. This extraction can be done for as many consecutive heartbeats as wanted. The labels are attributed by the central peak (whose index is the integer value of half the number of peaks). Once the windows are defined, we use interpolation to standardize the vectors, making them suitable for deep-learning purposes.

\paragraph{Feature Engineering.} Once those heartbeats are extracted, we build relative features. Literature \cite{Awais2017,Pyakillya2017,Blanco-Velasco2008,Luz2016} screening brought us to the discrete Fourier transform of each window, the linear relationships between each temporal components (P, Q, R, S, T), and the statistical values given by the extrema, mean, standard deviation, kurtosis, skewness, entropy, crossing-overs and PCA reduction to 10 components.

\begin{figure*}[hb!]
\centering
\includegraphics[width=0.95\textwidth]{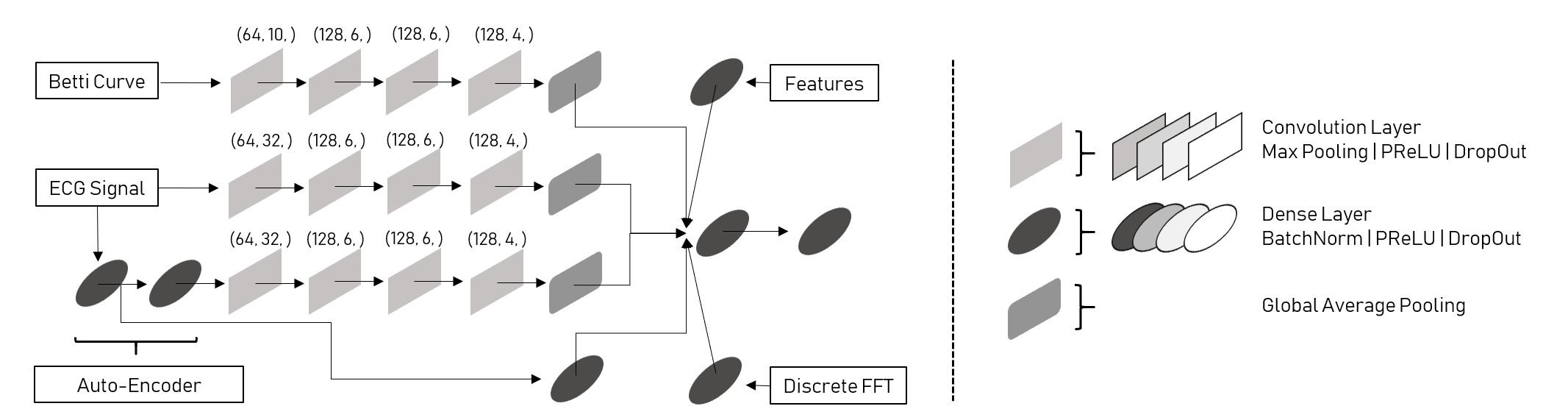}
\caption{General Overview of Deep-Learning Architecture}
\label{figure:architecture}
\end{figure*}

\subsection{Auto-Encoder}

Table \ref{table:dtb_description} gives a good overview of other issues we faced, such as an uneven distribution of labels and extreme minority classes. Furthermore, a challenging imbalance between normal and abnormal samples is noticeable, as it would be for any anomaly problem.

We decided to take advantage of the large amount of normal samples compared to abnormal samples, through unsupervised learning with auto-encoders \cite{autoencode}. The structure is made of six fully-connected hidden layers, developed in a symmetric fashion, with an input dimension of size 400 and a latent space of size 20. The loss is defined by the \textit{mean squared error}. The model is then trained on all the normal beats available, minimizing the reconstruction error between the input and the output. Once frozen, this model is integrated into our larger architecture. This reconstruction error, as presented in Figure \ref{figure:reconstruction_error}, is already a good indicator for anomaly detection, but not satisfactory for classification.

Such a constructor may be used in two different ways to deal with binary classification: by either using the encoded inputs as new features, or using the reconstruction error through a subtraction layer between the input signal and the reconstructed signal. Those solutions are respectively referred to by \textit{encoder} and \textit{auto-encoder} in our architectures. Another way of using this structure is to directly integrate it into the deep-learning model. The concurrent optimization of two models is thus necessary, building a relational encoding space relative to the task. This is the strategy that has been applied for multi-class classification.

\subsection{Architecture}

Once the preprocessing done, we undertook the construction of our deep-learning approach to deal with the multi-modality of inputs. Our first objective was to determine whether the heartbeats are normal or abnormal, before determining their classification. The aim of such strategy is to avoid the issue of great imbalance between normal and abnormal samples, while focusing on an easier task before multi-class classification. An overview of our architecture is given in Figure \ref{figure:architecture}.

\paragraph{Channels.} For the auto-encoder, we use a convolution channel to deal with the subtracting layer, and a fully-connected layer to deal with the feature map given by the latent space. The input signals and the Betti curves are fed in convolution channels, aiming at extracting the right patterns \cite{Kachuee,Xia2018,Isin2017,Rahhal2016,Clifford,Rajpurkar}. The other inputs (both features and discrete Fast Fourier Transform coefficients) are injected into fully connected networks.

\paragraph{Annealed DropOut.} As we launched a first battery of tests, we were confronted to the unexpected strong influence of the DropOut parameter. Its value could dramatically change the results. Since DropOut is of great help for generalization, we sought a way to deal with that issue. A solution came from the annealing dropout technique \cite{Rennie,JimmyBaBrendanFrey}, which consists in scheduling the decrease of the dropout rate. It helped us stabilizing the results.

\section{Experimental Results}

From the problem presentation, we highlighted two issues, relative to any healthcare machine-learning problem: imbalanced datasets and individual differences. The fewer the patients and the bigger the imbalance, the greater the influence of individual differences. To deal with the issue of imbalance, we introduced our auto-encoder architecture, while dealing with the individual differences by introducing Topological Data Analysis. Once we established our solution to both imbalance and individual differences, we aimed at developing our own approach and validation. As we mentioned earlier, two ways have been explored, both for performance enhancement and reduction of the influence of imbalance. The first one has been to detect whether a heartbeat is normal or abnormal, in order to get a first classification. The second one has been multi-class classification (13 classes) on the arrhythmic heartbeats only. Our objective is to introduce a new benchmark to attest that TDA (and auto-encoders) do improve generalization for arrhythmia classification.

\paragraph{Training Parameters.} Different methods have been used for the model training and optimization. Firstly, all the channels described previously are concatenated into one fully-connected network, dealing with all the obtained feature maps concurrently. Secondly, all the activation layers used are \textit{PReLU}, initialized with \textit{he\_normal} \cite{He,Srivastava}. Thirdly, the dropout has been parametrized according to the strategy of the \textit{annealing dropout}, from a rate of 0.5 to a rate of 0.0 after 100 epochs. Concerning the losses, we used the \textit{categorical\_crossentropy} or \textit{binary\_crossentropy} for the classification model, and \textit{mean\_squared\_error} for the auto-encoder structure. \textit{Adadelta} was used for optimization with an initial learning rate of \textit{1.0}.

\paragraph{Testing Methodology.} Dealing with a health issue, the testing methodology has to be rigorously defined to accurately analyze the performances. A great importance was given to generalization abilities of the developed models. For that purpose, our strategy aimed at performing \textbf{patient-based cross-validation}, which means that for each model, train and validation sets were build on a fraction of the available patients, while the remaining patients constituted the test set. This way, the validation score would demonstrate the ability of the model to dissociate arrhythmias on known patients, while the test score would demonstrate the ability of the model to detect arrhythmias on new patients. By using permutations of all the available patients, we were able to train, validate and test each model on all patients. The results presented in the following parts stem from a cross-validation keeping 5 unique patients for testing at each cross-validation permutation.

\subsection{Channel Comparison}

We first quantified the importance of each introduced channel, by turning them off and on. Such strategy allowed us to specifically quantify the influence of the introduced TDA channel, as improving the general ability of our architecture to both detect and classify arrhythmias. We tested our architecture through a patient-based cross-validation. Our format consisted in 10 experiments, made with the underlying will of generalization: \textbf{for each experiment}, 225 patients are used for both training (70\%) and validation (30\%), while 15 patients are kept for testing. For the purpose of validation, each subset of 15 patients is not overlapping between experiments. 

\begin{table}[ht!]
{\footnotesize {\centerline{\begin{tabular}{c|c|c|c|c}
\toprule
ID &
\multicolumn{2}{c}{Arrhythmia Detection} & \multicolumn{2}{c}{Arrhythmia Classification} \\
\midrule
& \textit{With TDA} & \textit{Without TDA} & \textit{With TDA} & \textit{Without TDA} \\
\midrule
0 & \textbf{0.99} & 0.98 & \textbf{0.73} & 0.68 \\
1 & \textbf{0.96} & 0.90 & \textbf{0.75} & 0.69 \\
2 & 0.85 & \textbf{0.86} & \textbf{0.68} & 0.65 \\
3 & 0.94 & \textbf{0.95} & 0.95 & \textbf{0.96} \\
4 & \textbf{0.85} & 0.80 & \textbf{0.97} & 0.97 \\
5 & \textbf{0.87} & 0.77 & \textbf{0.96} & 0.93 \\
6 & 0.78 & \textbf{0.80} & \textbf{0.94} & 0.93 \\
7 & \textbf{0.81} & 0.63 & \textbf{0.90} & 0.80 \\
8 & \textbf{0.79} & 0.65 & \textbf{0.85} & 0.78 \\
9 & 0.84 & \textbf{0.86} & \textbf{0.68} & 0.47 \\
\bottomrule
\end{tabular}}}}
\caption[LoF entry]{\centering Weighted Test Accuracy for Channel Comparisons. TDA improves both detection and classification, with a major improvement in arrhythmia classification, but also greatly enhances performances reliability.}
\label{table:test_accuracies}
\end{table}

Finally, a closer look at Table \ref{table:test_accuracies} supports the importance of TDA. Its role is emphasized for multi-class classification, with a general greater improvement of performances. With this right combination of channels, we aimed for testing through patient-based cross-validation for both binary and multi-class classification. \textbf{For the purpose of the demonstration, the scores are weighted in order to compensate for the general imbalance.} Moreover, multi-class classification is not biased by normal samples since they have been extracted beforehand. This finally supports the generalization role of TDA, that is expected to bring improvements combined with other deep-learning architectures as well.

\subsection{Arrhythmia Detection}

Our first benchmark dealt with arrhythmia detection (binary classification). It consisted in using our architecture, enhanced with the (auto-) encoder trained in an unsupervised manner on  normal beats. The model determined by channel comparison has thus been used for cross-validation. Each instance of cross-validation has been made by randomly undersampling the majority class to obtain balanced datasets. It takes approximately 10 hours to train on a GPU (GeForce GTX). We used the data structure previously presented to test over the 240 patients we have in our datasets. Moreover, to tackle the issue of anomaly detection, and accelerate the process of validation, each cross-validation round is respectively built out of a set of 5 unknown patients. The mean accuracy score is \textbf{98\%} for validation and \textbf{90\%} for test. This approach shows great generalization abilities. Unfortunately, no other paper do use those test settings for comparison. With a closer look on the results, the low performances appear on patients for which it was hard to recognize their normal beats. It also means that more patients may improve the generalization abilities of the model. However, its performances on the validation results prove its abilities to learn about specific patients (suitable for personalized devices issues).

\subsection{Arrhythmia Classification}

The same strategy has been applied for multi-class classification. The greatest channel influence is provided by TDA and the encoder. As a consequence, we reduced the original model to the one composed of four channels in the same fashion than we did for anomaly detection. Moreover, the influence of those channels is greater than observed for binary classification. Since the previous approach was not enough, we went further with 13-class classification. The models proved their ability to learn about heartbeat condition through cross-validation, with a mean validation score of \textbf{97.3\%}, while being able to generalize this acquired knowledge on patients it never saw, with a mean testing accuracy of \textbf{80.5\%}. Once again, literature does not provide comparable settings. The use of cross-validation focused on the generalization ability of the model. By also removing the normal beats, we focus on differences between the different arrhythmias, and remove the influence of imbalance that is generally found in the scores presented in the literature.

\section{Benchmarks Comparison}

\subsection{Premature Ventricular Heartbeats Detection}

Since those open-source datasets have been exploited by others, we sought to compare our own architecture to existing benchmarks. Our claim is an enhanced generalization thanks to TDA and the auto-encoder architecture we developed. To support it, our first confrontation has been made with \cite{Li2018}, which focuses on the detection of premature ventricular contractions (PVC). The detection of a specific arrhythmia is a particular case of anomaly detection (one-vs-all), for which our architecture is suitable. The results we obtained are presented in tables \ref{table:li_mit_bih} and \ref{table:li_all_physionet}, and support the generalization ability of our model. For the comparison, we applied the settings used in the paper. Out of the 48 initial patients in the MIT-BIH Arrhythmia Database, 4 patients are discarded. The remaining patients are split in two groups: 22 are used for training and validation, 22 are used for testing. The objective of such approach is to use the premature ventricular contractions, which are the majority class among the available arrhythmias. The results we obtained with such configuration are given in Table \ref{table:li_mit_bih} for the MIT-BIH Arrhythmia, where \textit{PPV} stands for \textit{Positive Predictive Value}.

\begin{table}[ht!]
{\footnotesize {\centerline{\begin{tabular}{c|c|c|c}
\toprule
Paper & \textbf{Accuracy} & \textbf{PPV} & \textbf{Sensitivity} \\
\midrule
Proposed & 99.2\% & 95.1\% & 95.5\% \\
Li et al. (2018) & 99.4\% & 93.9\% & 98.2\% \\
\bottomrule
\end{tabular}}}}
\caption{Detection of PVC - MIT-BIH Database}
\label{table:li_mit_bih}
\end{table}

However, they went further by considering the five databases. Our model being suited to that larger amount of data, we also compared our performances to their new settings. This time they split the group of 240 patients in two, both 120 for training and validation, and 120 for testing. Once again, our performances are given in Table \ref{table:li_all_physionet}. This experiment shows a great enhancement due to samples augmentation in this case of one-vs-all classification.

\begin{table}[ht!]
{\footnotesize {\centerline{\begin{tabular}{c|c|c|c}
\toprule
Paper & \textbf{Accuracy} & \textbf{PPV} & \textbf{Sensitivity} \\
\midrule
Proposed & 96.1\% & 97.0\% & 96.0\% \\
Li et al. (2018) & 95.6\% & 94.1\% & 92.7\% \\
\bottomrule
\end{tabular}}}}
\caption[LoF entry]{\centering Detection of PVC - Augmented Databases. With more patients, those settings imply a greater need for generalization, which emphasizes our model ability to learn the concept of PVC and apply it to unknown patients.}
\label{table:li_all_physionet}
\end{table}

\subsection{8-Classes Classification}

The previous experiment is a specific use-case for our architecture. In this second comparison, we focus on 8-classes classification \cite{Jun}. Once again, our claim is better generalization, which is done thanks to patient-based cross-validation.  Nonetheless, their settings imply a limitation to the MIT-BIH Arrhythmia Database, from which they select 8 classes, comprising normal beats. Unexpectedly, this selection does not correspond to the majority classes. Our performances are compared in Table \ref{table:jun_all_physionet}, extending the results they present in their paper. We pinpoint, thanks to those results, the generalization ability of our model, which has better positive predictive value (here precision) and sensitivity. It finally underlines a more efficient classification.

\begin{table}[ht!]
{\footnotesize {\centerline{\begin{tabular}{c|c|c|c|c}
\toprule
  Paper & Classes & \textbf{Acc} & \textbf{PPV} & \textbf{Sen} \\
\midrule
  Proposed & 8 & 99.0\% & 99.0\% & 98.5\% \\
  Jun et al. (2017) & 8 & 99.0\% & 98.5\% & 97.8\% \\
  Kiranyaz et al. (2017) & 5 & 96.4\% & 79.2\% & 68.8\% \\
  Güler et al. (2018) & 4 & 96.9\% & - & 96.3\% \\
  Melgani et al. (2008) & 6 & 91.7\% & - & 93.8\% \\
\bottomrule
\end{tabular}}}}
\caption{8-Classes Classification - MIT-BIH Database}
\label{table:jun_all_physionet}
\end{table}

\section{Conclusion}

We developed a new approach to deal with the issue of generalization in arrhythmia detection and classification. Our innovative architecture uses common source of information, Topological Data Analysis and auto-encoders. We supported our claim of improved generalization with scores reaching the performances of state of the art methods, and above. Our experiments pinpoint the strengths of TDA and auto-encoders to improve generalization results. Moreover, the modularity of such model allows us to build and add new channels, such as a possible channel based on the Wavelet transform \cite{Xia2018}, which also gives a good description of the ECG time-series. Finally, we give a new benchmark on five open-source datasets, and as it is often the case in deep-learning, we still envision greater performances with larger datasets such as \cite{Clifford}.

\end{document}